\documentclass[sigconf, screen]{acmart}

\usepackage{multirow}
\usepackage[dvipsnames]{xcolor}

\usepackage{tikz}
\usepackage{enumitem}
\usetikzlibrary{arrows.meta}
\AtBeginDocument{%
  }

\newcommand\blfootnote[1]{%
  \begingroup
  \renewcommand\thefootnote{}\footnote{#1}%
  \addtocounter{footnote}{-1}%
  \endgroup
}

\setcopyright{acmlicensed}
\copyrightyear{2026}
\acmYear{2026}
\acmDOI{XXXXXXX.XXXXXXX}
\acmConference[MM '26]{Proceedings of the 34th ACM International Conference on Multimedia}{10–14 November 2026,}{Rio de Janeiro, Brazil}
\acmBooktitle{Proceedings of the 34th ACM International Conference on Multimedia (MM '26), 10–14 November 2026, Rio de Janeiro, Brazil}

\begin{document}

\title{Learning Speaker Identity Beyond Language and Modality Constraints: Insights from the POLY-SIM 2026 Challenge}


\author{Marta Moscati$^{\ast}$}
\affiliation{%
  \institution{Institute of Computational Perception, Johannes Kepler University}  \institution{Albatross AI}
  \city{Linz}
  \country{Austria}}
\email{marta.moscati@jku.at}

\author{Muhammad Saad Saeed$^{\ast}$}
\affiliation{%
  \institution{University of Michigan-Flint}
  \city{Flint}
  \country{United States of America}}
\email{msaads@umich.edu}

\author{Marina Zanoni}
\affiliation{%
  \institution{BDO DIGITAL Gmbh}
  \city{Düsseldorf}
  \country{Germany}}
\email{marina.zanoni@bdo.de}

\author{Mubashir Noman}
\affiliation{%
  \institution{Mohamed bin Zayed University of Artificial Intelligence}
  \city{Abu Dhabi}
  \country{United Arab Emirates}}
\email{mubashir.noman@mbzuai.ac.ae}

\author{Rohan Kumar Das}
\affiliation{%
  \institution{Fortemedia}
  \city{Singapore}
  \country{Singapore}}
\email{rohankd@fortemedia.com}

\author{Monorama Swain}
\affiliation{%
\institution{Institute of Computational Perception, Johannes Kepler University}
  \city{Linz}
  \country{Austria}}
\email{monorama.swain@jku.at}

\author{Yassin Terraf}
\affiliation{%
  \institution{College of Computing, Mohammed VI Polytechnic University}
  \city{Benguerir}
  \country{Morocco}}
\email{yassin.terraf@um6p.ma}

\author{Yufang Hou}
\affiliation{%
  \institution{IT:U Interdisciplinary Transformation University}
  \city{Linz}
  \country{Austria}}
\email{yufang.hou@it-u.at}

\author{Elisabeth Andre}
\affiliation{%
  \institution{University of Augsburg}
  \city{Augsburg}
  \country{Germany}}
\email{elisabeth.andre@uni-a.de}

\author{Khalid Mahmood Malik}
\affiliation{%
  \institution{University of Michigan-Flint}
  \city{Flint}
  \country{United States of America}}
\email{drmalik@umich.edu}

\author{Markus Schedl}
\affiliation{%
  \institution{Institute of Computational Perception, Johannes Kepler University}
  \institution{Human-centered AI Group, AI Lab, Linz Institute of Technology, Austria}
  \city{Linz}
  \country{Austria}}
\email{markus.schedl@jku.at}

\author{Shah Nawaz$^{\ast \dagger}$}
\affiliation{%
  \institution{Institute of Computational Perception, Johannes Kepler University}
  \city{Linz}
  \country{Austria}}
\email{shah.nawaz@jku.at}


\renewcommand{\shortauthors}{Marta Moscati et al.}

\begin{abstract}
\blfootnote{\textsuperscript{$\ast$}Equal contribution.}
\blfootnote{\textsuperscript{$\dagger$}Corresponding author.}
Multimodal speaker identification systems typically assume the availability of complete and homogeneous audio–visual modalities during both training and testing, and assume each speaker only speaks a single language. However, in real-world applications, such assumptions often do not hold. Visual or audio information may be missing due to occlusions, camera or microphone failures, or privacy constraints. Multilingual speakers introduce additional complexity due to linguistic variability across languages. These situations constitute substantial challenges for the robustness and generalization capabilities of multimodal speaker identification systems. Aim of the POLY-SIM 2026  challenge is to address these aspects of speaker identification and to provide a standardized setup for the comparison of the proposed solutions.
\end{abstract}




\maketitle

\section{Introduction}
The face and voice of a person have unique characteristics and they are often used as biometric measures for speaker identification, either individually or combined together as a  multimodal input signal~\cite{jain2004introduction}.
Recent advancements in face-voice association have been fueled by the curation of large-scale audio-visual datasets such as VoxCeleb~\cite{nagrani2017voxceleb,chung18b_interspeech,nagrani2020voxceleb} and VoxBlink~\cite{lin24j_interspeech}. These datasets enable the development of multimodal models~\cite{tao20b_interspeech,tao2021someone,jiang23c_interspeech,praveen2025lavvit,terraf2024comisi} for speaker identification. 
However, a major limitation of existing models, even when trained on multimodal data, is that they require both the audio and the visual modalities to achieve good performance. These models therefore experience performance deterioration if evaluated when one of the modalities is not available. This situation, referred to as \textit{missing modality}, is a well-known challenge in multimodal learning across several tasks~\cite{ma2022multimodal,lin2023missmodal,saeed2024modality,guo2024multimodal,ganhor2024multimodal,ganhor2025single,liaqat2025multimodal,breiteneder2026robust}.
A separate challenge, specific to tasks such as speaker identification, is that of language shifts: when trained on one language (e.g., English) and evaluated on another (e.g., German), existing methods experience a performance deterioration, often attributed to the acoustic and phonetic differences between the train and evaluation languages~\cite{nawaz2021cross,saeed2024fame,moscati2025}.
To foster the development of solutions to these situations, we held the Polyglot Speaker Identification with Missing Modality (POLY-SIM) grand challenge, in which participants investigated multimodal learning under missing-modality and cross-lingual settings. To mimic a broad set of real-world situations, the challenge consisted of four different evaluation protocols, as illustrated in Figure~\ref{fig:protocols}. These protocols are designed to evaluate how well multimodal models leverage partial inputs and maintain strong cross-lingual  speaker identification generalization .

\section{The POLY-SIM Grand Challenge}
In this section we describe the task of the POLY-SIM grand challenge, as well as the training and evaluation data to be used in each protocol as summarized in Figure~\ref{fig:protocols}. 

\subsection{Task Overview}
Models developed for the POLY-SIM grand challenge are developed for face-voice association formulated as multimodal speaker classification~\cite{moscati2025}.

\definecolor{trainbg}{RGB}{213,232,252}
\definecolor{p3bg}{RGB}{210,238,210}
\definecolor{p4bg}{RGB}{254,243,219}
\definecolor{p5bg}{RGB}{248,224,240}
\definecolor{wavegreen}{RGB}{0,150,70}
\definecolor{wavepurple}{RGB}{130,50,180}
\definecolor{wavepink}{RGB}{230,70,130}
\definecolor{wavedark}{RGB}{30,30,100}
\definecolor{wavegold}{RGB}{200,140,20}

\newcommand{\waveformEN}[3]{%
  \foreach [count=\i] \h in {
    0.06, 0.10, 0.18, 0.28, 0.36, 0.24, 0.38, 0.44, 0.28, 0.20,
    0.36, 0.40, 0.26, 0.16, 0.30, 0.20, 0.14, 0.18, 0.09, 0.05%
  }{%
    \draw[#3, line width=1.3pt]
      ({#1 + (\i - 10.5) * 0.075}, {#2 - \h/2}) --
      ({#1 + (\i - 10.5) * 0.075}, {#2 + \h/2});
  }%
}

\newcommand{\waveformUR}[3]{%
  \foreach [count=\i] \h in {
    0.08, 0.32, 0.09, 0.38, 0.08, 0.40, 0.09, 0.36, 0.08, 0.39,
    0.09, 0.37, 0.08, 0.38, 0.09, 0.35, 0.08, 0.30, 0.09, 0.08%
  }{%
    \draw[#3, line width=1.3pt]
      ({#1 + (\i - 10.5) * 0.075}, {#2 - \h/2}) --
      ({#1 + (\i - 10.5) * 0.075}, {#2 + \h/2});
  }%
}

\newcommand{\personicon}[3]{%
  \draw[#3, fill=#3!15, rounded corners=5pt, line width=1pt]
    (#1-0.55, #2-0.55) rectangle (#1+0.55, #2+0.55);
  \draw[#3, fill=#3!30, line width=1.2pt] (#1, #2+0.22) circle (0.18);
  \draw[#3, line width=1.2pt]
    (#1-0.28, #2-0.3)
    to[out=90, in=180] (#1, #2+0.02)
    to[out=0,  in=90] (#1+0.28, #2-0.3);
}

\newcommand{\missingperson}[2]{%
  \draw[gray!70, dashed, fill=gray!10, rounded corners=5pt, line width=1pt]
    (#1-0.55, #2-0.55) rectangle (#1+0.55, #2+0.55);
  \draw[gray!70, dashed, line width=1.2pt] (#1, #2+0.22) circle (0.18);
  \draw[gray!70, dashed, line width=1.2pt]
    (#1-0.28, #2-0.3)
    to[out=90, in=210] (#1, #2+0.02)
    to[out=30,  in=90] (#1+0.28, #2-0.3);
  \draw[red!80, line width=1.5pt] (#1, #2) circle (0.42);
  \draw[red!80, line width=1.5pt] (#1-0.30, #2-0.30) -- (#1+0.30, #2+0.30);
}

\newcommand{\facephoto}[2]{%
  \draw[gray!60, fill=gray!10, rounded corners=4pt, line width=0.8pt]
    (#1-0.55, #2-0.7) rectangle (#1+0.55, #2+0.7);
  \draw[gray!60, line width=1.2pt] (#1, #2+0.22) circle (0.18);
  \draw[gray!60, line width=1.2pt]
    (#1-0.28, #2-0.3)
    to[out=90, in=180] (#1, #2+0.02)
    to[out=0,  in=90] (#1+0.28, #2-0.3);
}

\begin{figure}
\centering
\resizebox{\columnwidth}{!}{%
\begin{tikzpicture}[font=\sffamily\small]

\filldraw[fill=trainbg, draw=gray!50, rounded corners=10pt, line width=0.8pt]
  (-0.2, 4.7) rectangle (9.8, 7.5);
\node[font=\sffamily\bfseries\normalsize] at (4.8, 7.2) {Train};

\facephoto{1.0}{6.1}
\node[below, font=\sffamily\footnotesize] at (1.0, 5.4) {Face};
\node[font=\large] at (2.3, 6.1) {$+$};
\waveformEN{4.0}{6.1}{wavepurple}
\node[below, font=\sffamily\footnotesize] at (4.0, 5.65) {Voice (English)};
\draw[->, >=Stealth, thick] (5.4, 6.1) -- (6.2, 6.1);
\node[draw, fill=blue!12, rounded corners=5pt, line width=0.8pt,
      minimum width=1.5cm, minimum height=0.9cm, font=\sffamily\small]
      at (7.1, 6.1) {Model};
\draw[->, >=Stealth, thick] (7.95, 6.1) -- (8.5, 6.1);
\personicon{9.1}{6.1}{blue}

\filldraw[fill=p3bg, draw=gray!50, rounded corners=8pt, line width=0.8pt]
  (-0.2, 1.9) rectangle (4.8, 4.4);
\node[font=\sffamily\bfseries\footnotesize, align=center] at (2.3, 4.1)
  {P3:Inference (In-\\Language Multimodal)};
\facephoto{0.8}{3.15}
\node[below, font=\sffamily\footnotesize] at (0.8, 2.45) {Face};
\node[font=\large] at (2.0, 3.15) {$+$};
\waveformEN{3.3}{3.15}{wavepurple}
\node[below, font=\sffamily\footnotesize] at (3.3, 2.7) {Voice (English)};

\filldraw[fill=p4bg, draw=gray!50, rounded corners=8pt, line width=0.8pt]
  (5.2, 1.9) rectangle (9.8, 4.4);
\node[font=\sffamily\bfseries\footnotesize, align=center] at (7.5, 4.1)
  {P4:Inference (In-\\Language Missing-Face)};
\missingperson{6.3}{3.15}
\node[font=\large] at (7.3, 3.15) {$+$};
\waveformEN{8.6}{3.15}{wavepurple}
\node[below, font=\sffamily\footnotesize] at (8.6, 2.7) {Voice (English)};

\filldraw[fill=p5bg, draw=gray!50, rounded corners=8pt, line width=0.8pt]
  (-0.2, -0.9) rectangle (4.8, 1.6);
\node[font=\sffamily\bfseries\footnotesize, align=center] at (2.3, 1.3)
  {P5:Inference (Cross-\\Lingual Multimodal)};
\facephoto{0.8}{0.35}
\node[below, font=\sffamily\footnotesize] at (0.8, -0.35) {Face};
\node[font=\large] at (2.0, 0.35) {$+$};
\waveformUR{3.3}{0.35}{wavegold}
\node[below, font=\sffamily\footnotesize] at (3.3, -0.1) {Voice (Urdu)};

\filldraw[fill=p4bg, draw=gray!50, rounded corners=8pt, line width=0.8pt]
  (5.2, -0.9) rectangle (9.8, 1.6);
\node[font=\sffamily\bfseries\footnotesize, align=center] at (7.5, 1.3)
  {P6:Inference (Cross-\\Lingual Missing-Face)};
\missingperson{6.3}{0.35}
\node[font=\large] at (7.3, 0.35) {$+$};
\waveformUR{8.6}{0.35}{wavegold}
\node[below, font=\sffamily\footnotesize] at (8.6, -0.1) {Voice (Urdu)};

\end{tikzpicture}%
}
\caption{POLY-SIM 2026 evaluation protocol. The top blue box represents the training phase, for which English and complete (both audio and image) data is used. Each of the boxes below represents a different evaluation protocol (P).}
\label{fig:protocols}
\end{figure}
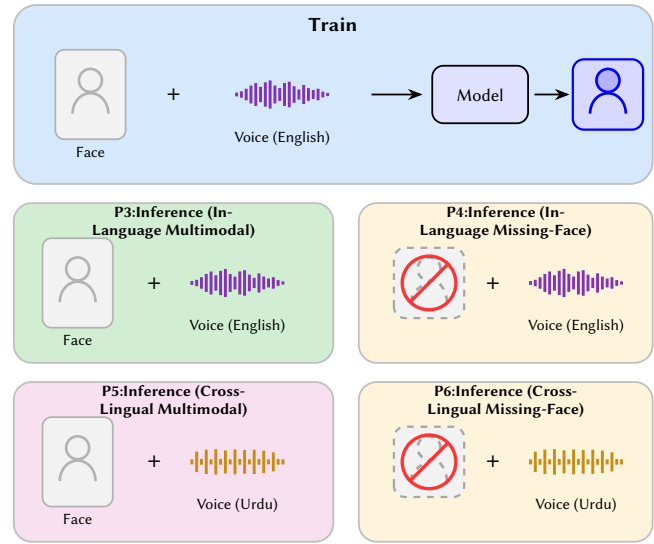

\subsection{Evaluation Protocol}
\label{sec:protocols}
Training data consist of paired samples of face images and speech segments. All train speech segments belong to the same language  (English) for all speakers.
At test time, the face modality might be missing, and the available speech segment might be in another language (Urdu). Combinations of these settings result in the different evaluation protocols (P) designed for the challenge and simulate real-world scenarios including both (i) the missing visual modality and (ii) the cross-lingual scenario.
\begin{enumerate}[label=P\arabic*., start=3]
    \item \textbf{In-language multimodal.} Train and test data consist of speakers speaking the same language (English), and both modalities are available at test time.
    \item \textbf{Missing-modality.} Train and test data consist of speakers speaking the same language (English). Test data consist of audio only, while faces are missing.
    \item \textbf{Cross-lingual and multimodal.} Train data consist of speakers speaking one language (English), while test data consist of the same speakers speaking a different language (Urdu). Both modalities are available at test time.
    \item \textbf{Cross-lingual and missing-modality.} Train data consist of speakers speaking one language (English), while test data consist of the same speakers speaking a different language (Urdu).  Test data consist of audio only, while faces are missing.
\end{enumerate}

\subsection{Dataset}
\begin{figure}[t]
    \centering
    \includegraphics[width=0.95\linewidth]{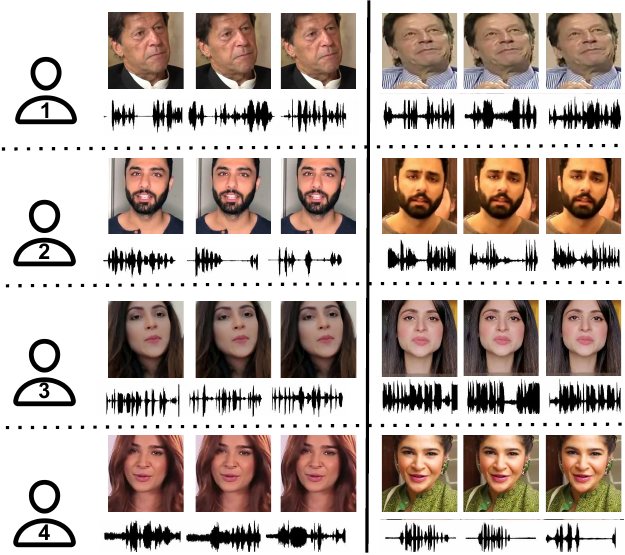}
    \caption{Audio-visual samples randomly selected from the MAV-Celeb dataset~\cite{nawaz2021cross,saeed2024synopsis,moscati2025linking}. The visual data contains different variations such as pose, lighting condition, and motion. (Left) The block shows data of speakers speaking English. (Right) The block shows data of the same speakers speaking Urdu language.}
    \label{fig:mavceleb}
\end{figure}

The grand challenge is based on the MAV-Celeb dataset~\cite{nawaz2021cross,saeed2024fame,moscati2025linking}. It consists of audio-visual samples of several speakers while speaking in interviews, talk shows, and television debates. These audio-visual samples were obtained from YouTube videos. 
Most importantly for what concerns POLY-SIM, for each speaker, the dataset consists of samples in which the speakers speaks two distinct languages (but only one language in each sample). For the challenge we selected the dataset subset consisting of speakers appearing while speaking English and Urdu. 
We adapted the dataset to the multimodal speaker identification task under missing-modality and cross-lingual scenarios by separating train and test data according to the protocols described in Section~\ref{sec:protocols}. 
Table~\ref{tab:dataset_stats} provides the statistics of the dataset, while Figure~\ref{fig:mavceleb} shows audio-visual samples from the dataset.
The dataset is publicly available and provides raw audio and image as well as representations encoded with state-of-the-art pretrained architectures.

\subsection{Baseline Method \& Starter Kit}
To allow participants to compare their approaches with existing methods, we released a pretrained instance of a recent multimodal method for face-voice association~\cite{saeed2022fusion}. 
The model consists of a two-branch architecture, in which each branch takes as input the embeddings of one of the two modalities (one branch for faces and one branch for voices). The embeddings to be used as input to the face-encoding branch are obtained using a well-established  convolutional neural network pretrained for facial recognition on a large-scale dataset~\cite{schroff2015facenet}.  The embeddings to be used as input to the voice-encoding branch are obtained using a well-established audio encoding network for speaker recognition~\cite{desplanques20_interspeech} and pretrained using the training data of the challenge, consisting only of one language (English). The multimodal model further combines the outputs of the face and voice branches by means of a fusion module. The architecture is optimized by means cross-entropy on the multimodal embeddings of different speakers and is therefore named  Fusion and Orthogonal Projection (FOP).
We refer the readers to Saeed et al.~\cite{saeed2022fusion} and to the repository of the grand challenge for more information on baseline method\footnote{\href{https://github.com/msaadsaeed/polysim}{https://github.com/msaadsaeed/polysim}}.

\begin{table}
\caption{MAV-Celeb dataset statistics for English--Urdu language pair.}
\centering
\renewcommand{\arraystretch}{1.05}
\setlength{\tabcolsep}{10pt}
\scalebox{0.85}{
\begin{tabular}{cccc}
\hline
\multirow{2}{*}{\textbf{Lang. Pair}} & \multirow{2}{*}{\textbf{Lang.}} & \textbf{Total Videos} & \textbf{Samples} \\
 &  & \textbf{(Tr./Dev./Eval.)} & \textbf{(Tr./Dev./Eval.)} \\
\hline
\multirow{2}{*}{English--Urdu}
 & English & 262 / 70 / 70 & 4039 / 1290 / 1521 \\
 & Urdu    & 415 / 70 / 70 & 9304 / 1779 / 1623 \\
\hline
\end{tabular}
}
\label{tab:dataset_stats}
\end{table}

\begin{figure*}[t]
    \centering
    \includegraphics[width=1.05\linewidth]{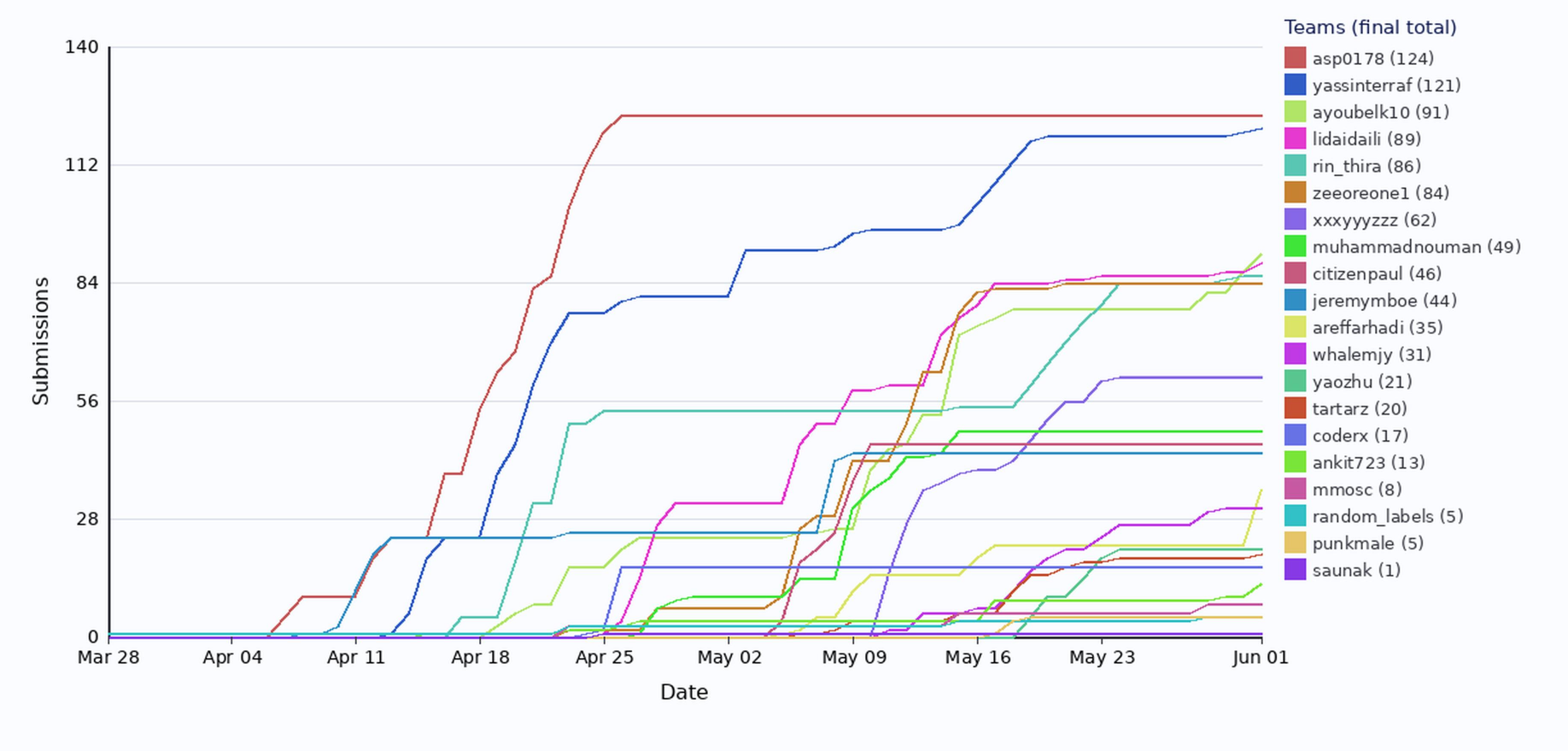}
    
    \caption{Cumulative submission counts per team over the grand challenge duration (March 27 -June 1, 2026). Each line represents the running total of submissions for one of the 18 participating teams, ordered by final cumulative total (highest to lowest). Steeper slopes indicate periods of higher submission activity, while plateaus indicate inactivity. A total of 952 submissions were recorded across all teams during the observation window.}
    \label{fig:daily_submissions}
    
\end{figure*}

\subsection{Evaluation Metric}
We evaluate models' performance in terms of accuracy on the task of speaker classification. Accuracy is measured in terms of $P$-accuracy, which measures the proportion of test pairs for which the system correctly predicted the matching identity among a set of $P$ candidates. $P$-accuracy is computed for each of the protocols separately. The overall score is the mean $P$-accuracy over all four protocols.

\subsection{Evaluation Plan} 
Comprehensive information regarding the grand challenge, including the challenge timeline, baseline methodology, evaluation protocol and metrics, submission system, and participation rules, is provided in the challenge evaluation plan provided to the participants when the challenge was announced,~\cite{moscati2026poly}.



\begin{table*}[!t]
\centering
\caption{Performance in accuracy (\%) for English--Urdu cross-lingual experiments.}
\renewcommand{\arraystretch}{1.15}
\setlength{\tabcolsep}{14pt}
\scalebox{0.85}{
\begin{tabular}{llccccc}
\toprule
\multirow{2}{*}{\textbf{Configuration}} & \multirow{2}{*}{\textbf{Method}} &
\textbf{P3} & \textbf{P4} & \textbf{P5} & \textbf{P6} & \multirow{2}{*}{\textbf{Avg.}} \\
\cline{3-6}
& & \multicolumn{1}{c}{Face--Audio (Eng.)} & \multicolumn{1}{c}{ Audio (Eng.)} &
\multicolumn{1}{c}{Face--Audio (Urdu)}     & \multicolumn{1}{c}{ Audio (Urdu)} & \\
\midrule


\multirow{4}{*}{Face-Audio (Eng.)} & Baseline FOP~\cite{saeed2022fusion} & 98.82 & 52.53  & 98.27 & 43.87 & 73.37 \\          
& MaskedFOP~\cite{elkhouzari2026maskedfop}  & 99.80 &  99.80 & 100   & 99.94 & 99.89 \\
& MRAF~\cite{jia2026missing}                & 100   & 98.48  & 100   & 99.32 & 99.56 \\
& AMR~\cite{zuo2026amr}                     & 99.93 & 97.50  & 100   & 98.83 & 99.07 \\


\bottomrule
\end{tabular}
}
\label{tab:v1_multimodal_configs}
\end{table*}

\subsection{Challenge Timeline and Results} 
The challenge consisted of two phases. For both phases, participants were allowed to use only the labels of the English training. In the first phase (March $27$, $2026$, to May $17$, $2026$) participants' approaches were evaluated on a  development set, whose labels were also available to participants. Each team was allowed to submit $15$ submissions per day, with a maximum number of $150$ submissions during the entire phase. This phase was meant to give participants the required environment to develop their systems and did not count towards the final challenge results. In the second phase (May $29$, $2026$ to June $01$, $2026$) participants' approaches were evaluated on an unseen evaluation set. Each team was allowed for a maximum of $15$ submissions over the whole evaluation phase.
Figure \ref{fig:daily_submissions} shows the cumulative submission counts per team over the whole grand challenge duration
. 
Table~\ref{tab:v1_multimodal_configs} presents the performance of the top-3 teams according to the averaged $P$-accuracy in the second evaluation phase, as well as the performance of the  baseline method FOP~\cite{saeed2022fusion}. 
The participating teams demonstrated a notable improvement over the baseline method, with the leading team, MaskedFOP~\cite{elkhouzari2026maskedfop}, achieving an accuracy of $99.89$\%, a considerable gain from the baseline accuracy of $73.37$\%. 
This substantial performance gap underscores the success of the challenge, which led to the development of effective methodologies that advanced the state-of-the-art for multimodal speaker identification under missing modality and cross-lingual settings. We attribute the performance gain of all teams over the baseline to the following consideration. 
Although none of the teams used test ground-truth labels (i.e., speakers' identities) for training the proposed methods, a situation that would constitute a breach of the challenge rules, successful approaches utilized test input data to develop unsupervised approaches  (e.g., for clustering). In contrast, the baseline method does not utilize test input features in any way, other than as input at inference time. 

\section{Lessons Learned as Organizers}
From an organizational perspective, we identified several important lessons that may inform the design and execution of future grand challenges.
\begin{enumerate}[label=\arabic*.]
    \item During the challenge, organizers observed that some participants submitted identical prediction files for the paired settings P3/P4 and P5/P6. Since P3 and P5 correspond to multimodal evaluation, while P4 and P6 correspond to missing-modality and cross-lingual evaluation, identical submissions undermine the intended purpose of the benchmark. Therefore, we introduced an additional validation mechanism to ensure that submissions for multimodal, missing-modality and cross-lingual tracks remained distinct.   
    \item While the top-performing approaches successfully handled the grand challenge and achieved impressively accurate predictions, most of them exploited the availability of test-time input data from both source and target languages, i.e., English and Urdu, to develop unsupervised adaptation strategies, such as clustering-based methods. However, assuming that enough representative samples in both languages are available is restrictive in real-world scenarios, where one of the languages is typically less represented, and is hence treated as unheard at inference time. Therefore, future research should focus on cross-lingual multimodal methods that generalize to unheard target languages without relying on test-time input data for generalization.
\end{enumerate}

\section{Acknowledgments}
This research was funded in whole or in part by the Austrian Science Fund (FWF): Cluster of Excellence \href{https://www.bilateral-ai.net/home}{\textcolor{blue}{\textit{Bilateral Artificial Intelligence}}} (\url{https://doi.org/10.55776/COE12}), the doc.funds.connect project \href{https://dfc.hcai.at/}{\textcolor{blue}{\textit{Human-Centered Artificial Intelligence}}} (\url{https://doi.org/10.55776/DFH23}), and the PI project \href{https://doi.org/10.55776/P36413}{\textcolor{blue}{\textit{Intent-aware Music Recommender Systems}}} (\url{https://doi.org/10.55776/P36413}).
For open access purposes, the authors have applied a CC BY public copyright license to any author-accepted manuscript version arising from this submission.


\balance
\bibliographystyle{ACM-Reference-Format}
\bibliography{sample-base}

\end{document}